\documentclass{article}

\usepackage{fullpage}
\usepackage[small]{caption}
\usepackage{fancyhdr}
\usepackage{mathpazo}
\usepackage{natbib}
\usepackage{titling}

\usepackage[utf8]{inputenc} 
\usepackage[T1]{fontenc}    
\usepackage{hyperref}       
\usepackage{url}            
\usepackage{booktabs}       
\usepackage{amsmath}
\usepackage{amssymb}
\usepackage{amsfonts}       
\usepackage{nicefrac}       
\usepackage{microtype}      
\usepackage{xcolor}         
\usepackage{graphicx}
\usepackage{float}
\usepackage{wrapfig}
\usepackage{subcaption}
\usepackage{algorithm}
\usepackage{algorithmic}
\usepackage{inconsolata}

\title{Explaining Attention with Program Synthesis}

\date{}

\author{%
\begin{minipage}{0.4\textwidth}
\centering
  Amiri Hayes\thanks{Preprint. Correspondence to: \texttt{akh5@njit.edu}. Work performed at MIT.} \\
  NJIT
  \end{minipage}
  \begin{minipage}{0.5\textwidth}
  \centering
  Belinda Z. Li ~~~~ 
  Jacob Andreas \\
  MIT EECS 
  \end{minipage}
}
\begin{document}
\maketitle

\begin{abstract}
A longstanding goal of research on interpretable deep learning is to replace opaque neural computations with human-meaningful symbolic descriptions. In this paper, we propose an approach for approximating the behavior of components of deep networks with executable programs. We focus on attention heads in transformer language models. For a given head, we first compute its associated attention matrices on a collection of randomly selected training examples. Next, we prompt a pre-trained language model with a summary of these matrices, and instruct it to generate a set of Python programs that can reproduce the associated attention patterns given only text from the input sentence. Finally, we re-rank programs  according to how well our final set of programs predict behavior on held-out inputs.  We demonstrate that a set of fewer than 1,000 such generated programs can reproduce the attention patterns of heads in GPT-2, TinyLlama-1.1B, and Llama-3B, achieving a median Intersection-over-Union similarity as high as 70\% on TinyStories.
Moreover, the best-fit programs can \emph{replace} neural attention heads without substantially affecting model behavior: replacing 25\% of attention heads with programmatic surrogates across the three models incurs as low as a 16\% average perplexity increase, while maintaining performance on a variety of downstream question answering benchmarks. This work contributes a scalable pipeline for reverse-engineering attention heads in transformer models using human-readable, executable code, advancing a path toward symbolic transparency in neural models.



\end{abstract}
\section{Introduction}

Understanding the computations performed by deep networks in algorithmic terms is a long-standing problem in machine learning \cite{erhan2009visualizing, zeiler2014visualizing, Nanda2023}. Past work approaching this problem has attempted to assign meaning to neurons or other distributed features in a top-down way (by training probes for human-defined concepts of interest \cite{HewittManning2019, Tenney2019}), or in a bottom-up way (by generating summaries of the inputs that activate a feature, or the outputs that the feature promotes \cite{Mu2020, HernandezEtAl2022, Bills2023}).
Much of the work in the later category associated features with natural language descriptions \cite{Bricken2023, Cunningham2023}, whose interpretations may themselves be ambiguous or hard to formalize.
While powerful, these methods stop short of providing a full, formal description of neural computation. 

In this paper, we develop an alternative approach to deep network interpretability based on program synthesis. To explain the computation performed by some component of a deep network, we search for a piece of executable code that approximates the computation performed by that component. Executable programs occupy a natural middle ground between the complexity of billion-parameter models and natural language explanations; they offer a medium that is both human-readable and formally verifiable \cite{Bills2023}. Unlike natural language descriptions, programs can be directly substituted for neural components, enabling causal validation of inferred explanations and opening a path toward model editing grounded in symbolic computation.

\begin{figure}[t]
  \begin{center}
    \centerline{\includegraphics[width=\columnwidth]{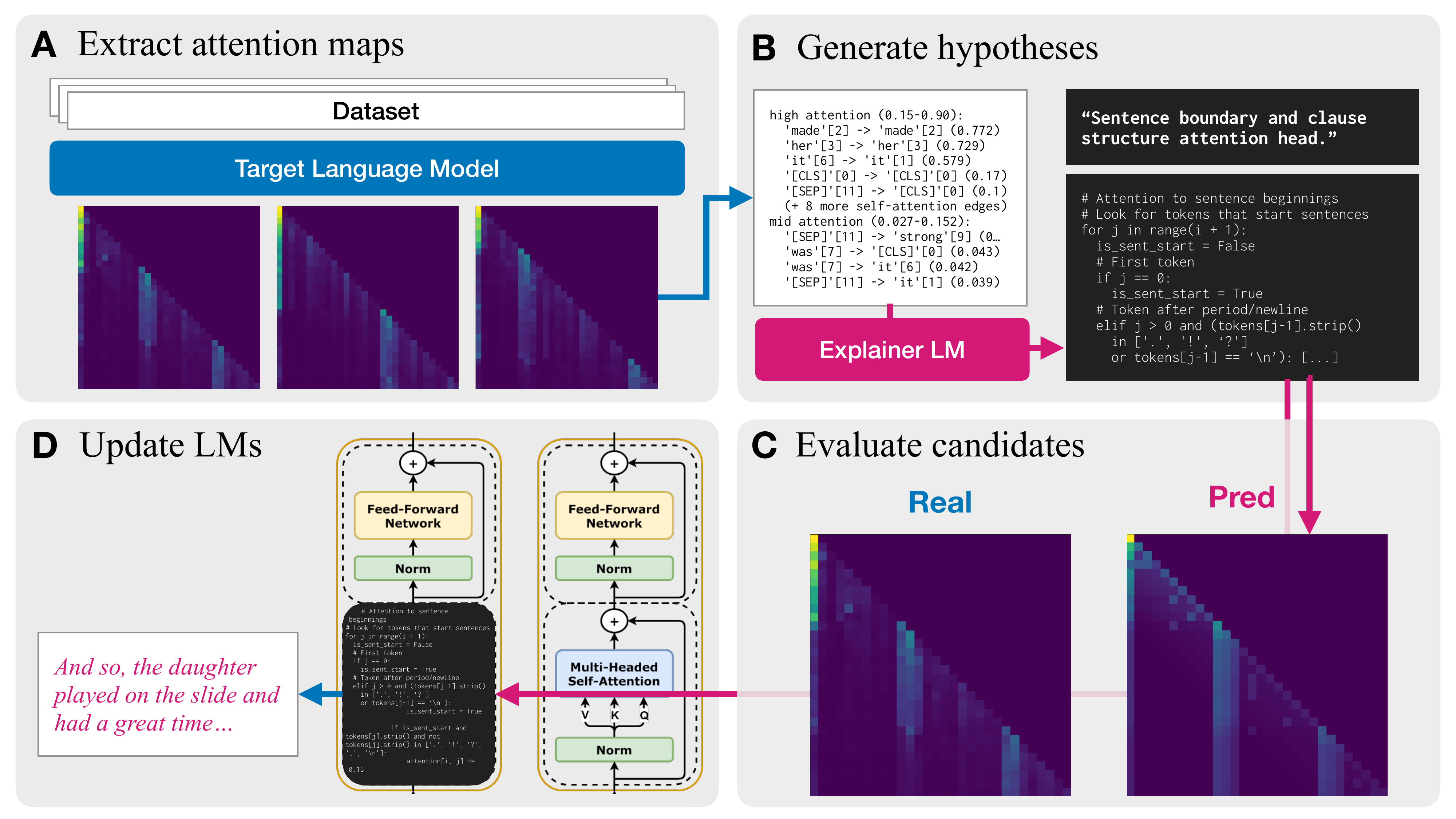}}
    \caption{
      Synthesizing programmatic representations of attention heads in transformer models. Clockwise from top left: \textbf{A)} Given a target model for explanation, we extract its attention pattern on a collection of representative inputs. \textbf{B)} We summarize these attention patterns in a textual prompt, and instruct an explainer language model to synthesize one or more candidate Python programs that can reproduce attention patterns given only input text. \textbf{C)} Candidate programs are compared to original attention patterns, and the highest-scoring ones are selected. \textbf{D)} These programs can be directly inserted into models, replacing learned attention heads with symbolic programs.
    }
    \label{description_of_approach}
  \end{center}
\vspace{-2em}
\end{figure}

We focus on generating programmatic explanations of attention heads in transformer language models (LMs) \cite{Vaswani2017}. For each attention head in a model of interest, we extract a set of example attention maps on training examples, then prompt another LM to generate a set of candidate Python programs that can reproduce those attention maps given only input text (Figure~\ref{description_of_approach}). Finally, we re-rank this collection of programs for each head in a model to obtain a best-fit program for each attention head. Across BERT-Base, GPT-2-Small, TinyLlama-1.1B and Llama-3B language models, we find that a substantial fraction of attention heads can be approximated with executable programs to a high degree of accuracy. We test whether the programs we find recover their associated heads' causal function by replacing heads with programs. 

Our results show that, even in state-of-the-art language models, a substantial fraction of attention patterns can be understood in symbolic terms; so much so that we can actually replace trained neural components with symbolic surrogates via executable code without substantially changing model behavior. More generally, they highlight the utility of modern (LM-driven) program synthesis methods as an alternative framework for approaching the broader question of how deep networks operate.

\section{Approach}

\subsection{Method Overview}

\newcommand{\mymodel}{\mathcal{M}} 

\newcommand{\myattention}{\mathcal{A}}
\newcommand{\mysynthesizer}{\mathcal{S}}

We seek to approximate the behavior of a \textbf{transformer model $\mymodel$}, which contains (among other modules) a set of \textbf{attention heads $\myattention_i$} in each layer. For a given \textbf{input sequence of tokens} $X$, each head $\myattention_i$ within $\mymodel$ projects the last layer's hidden representations $H$ into queries ($Q$) and keys ($K$) 
to produce an \textbf{attention matrix}:
\begin{equation}
    A = \mathrm{softmax}\left(\frac{QK^\top}{\sqrt{d_{k}}}\right)
\end{equation}
where $A \in \mathbb{R}^{n \times n}$ represents the directed weights between all token pairs in a sequence of length $n$. 
These attention weights then determine how much each token in a previous layer affect the token position of the current layer.
Given $\mymodel$, we wish to find for each $\myattention_i$ \textbf{some symbolic program $\pi$} that can map directly from $X \mapsto A$. Our framework, illustrated in Figure \ref{description_of_approach}, proceeds in two steps:

\paragraph{Attention Map Extraction} We first record the ground-truth attention activation matrices of $\mathcal{M}$ across a corpus of sequences processed by the target model $\mymodel$. This provides a set of empirical attention matrices $\{A_1, A_2, \dots, A_n\}$ that serve as the targets for our programmatic approximations.

\paragraph{Program Synthesis and Refinement} We define a space of symbolic programs $\Pi$, where each $\pi \in \Pi$ is an executable Python function that takes $X$ as input and outputs a hypothesized attention matrix $\hat{A} = \pi(X)$. We utilize an interactive program synthesis agent $\mysynthesizer$ (another LM) to generate these programs based on the patterns observed in the extracted maps. During program synthesis and refinement, we rank and select candidate programs by their Intersection-over-Union (IoU) agreement with the target attention patterns:
\begin{equation}
    \label{eq:iou}
    \mathrm{IoU}(A, \hat{A}) = \frac{\sum_{i,j} \min(A_{i,j},
    \hat{A}_{i,j})}{\sum_{i,j} \max(A_{i,j}, \hat{A}_{i,j})}
\end{equation}
The top candidate undergoes one round of feedback-conditioned refinement. In the remainder of this section, we describe the main subtasks in the program synthesis step in more detail.

\subsection{Generating Functional Proxies via Program Synthesis}

\begin{figure}[t!]
\includegraphics[width=\columnwidth]{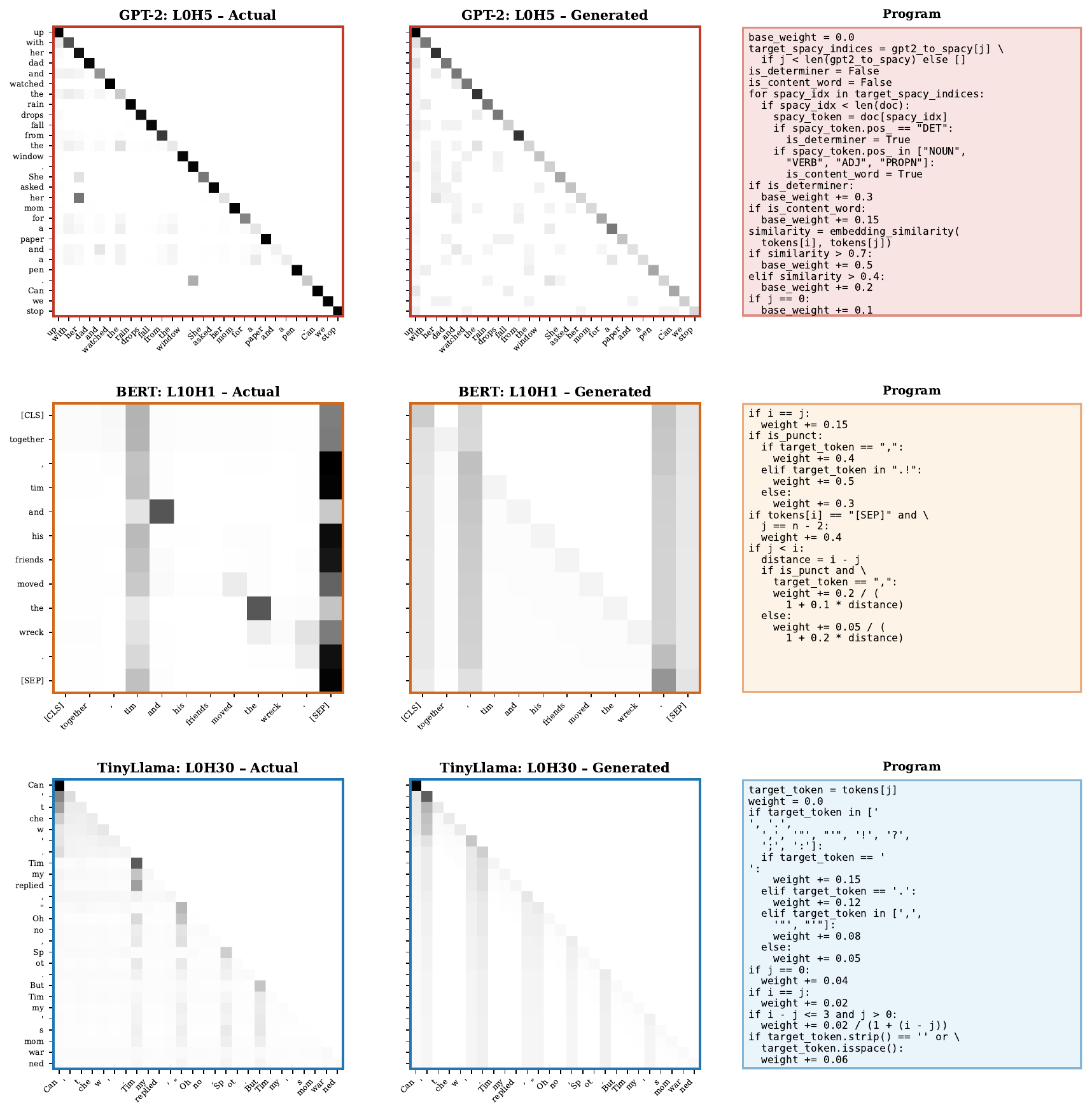}
\caption{Three attention heads in GPT2, TinyLlama and BERT models, their synthesized replacements, and (excerpts from) the associated programs. L$x$H$y$ means ``layer $x$, head $y$''. Programs often reproduce general attention patterns but sometimes hallucinate structural features (e.g.\ the diagonal attention in BERT).}
\end{figure}

To search in the space of symbolic programs $\Pi$, we utilize an auxiliary large language model
as a synthesis agent, denoted $\mathcal{S}$. Given a set of target attention matrices $A$ and the corresponding input sequences $X$, $\mathcal{S}$ is prompted to produce an executable candidate Python function $\pi$ such that each $\pi(X) \approx A$. Candidate functions $\pi$ are validated for syntax and executability, then scored against real attention using the Intersection-over-Union metric of Eq.~\ref{eq:iou}. Non-well-formed functions $\pi$ are assigned a score of zero. We then \textit{refine} our prediction by iterating through a sample of inputs $X$ to identify representative best and worst-scoring examples (with highest and lowest IoU), constructing structured error feedback by contrasting real and predicted attention patterns, and prompting $\mathcal{S}$ to produce a revised program $\pi'$. The refined program is re-validated and re-scored under the same similarity objective. The optimal proxy $\pi^*$ is then selected from $\{\pi, \pi'\}$ by maximizing similarity over held-out data sequences $(X', A') \in \mathcal{D}_{\text{val}}$.

Concretely, our pipeline first summarizes attention patterns, filtering for the top 2.5\% of attention
weights by magnitude to isolate the most salient token-pair interactions. These filtered patterns are formatted as token-pair weight summaries and embedded into structured prompts (approximately 4,000 tokens in length; full prompt template in Appendix~\ref{sec:appendix_prompt}). The synthesis agent $\mathcal{S}$ is provided access to \texttt{NumPy}, \texttt{spaCy} and \texttt{NLTK} for numerical and linguistic processing. The full program library used across all four models and used for all results in this paper was produced from fewer than 4,000 candidates using Claude Sonnet 4 at approximately \$150 in total API cost (roughly 35 million input tokens and 3.5 million output tokens). The final set of programs is composed of one program for each head in the four models, or 1,664 programs total.

\subsection{Evaluation Details}
\paragraph{Evaluation Metrics}
We evaluate our programs with two complementary metrics: a correlative metric that measures how closely synthetic programs reproduce attention patterns (attention alignment), and a causal metric that tests whether replacing real heads with our synthetic programs preserves model behavior (causal head replacement).

\textit{Attention alignment:}
Following past work \citep{bau2017network}, we measure the \emph{intrinsic} quality
of a synthesized program by the Intersection-over-Union (IoU) between its predicted
attention matrix and the real one. This is the same metric used to rank candidates
during synthesis, so selection and evaluation are scored consistently:

\textit{Causal head replacement:}
To validate that our symbolic programs are \textit{causally} faithful, we also perform interchange interventions \cite{Geiger2021}. We replace the neural attention matrix $A$ with the programmatic output $\pi(X)$ during the model's forward pass, and measure the resulting effect on downstream task performance. This tells us whether the program captures the attention-level features needed to maintain functional performance. Because interventions are costly, we use the IoU alignment metric above as a filter, attempting replacement only for heads with high-similarity symbolic programs. By measuring the resulting change in perplexity and task accuracy, we validate whether the proxy captures the necessary attention-level features to maintain the model's functional performance.

\paragraph{Datasets}

First, we \textbf{generate} programmatic attention patterns on TinyStories \cite{EldanLi2023}, which was selected for its relative simplicity: using simple, structured data simplifies the act of isolating specific head behaviors without the stochastic noise inherent in complex corpora \cite{Voita2019}. No preprocessing or sequence truncation was performed. We also \textbf{evaluate attention alignment} between our generated programs and true attention on held-out subset of TinyStories \citep{EldanLi2023} (\S\ref{sec:alignment_eval}). Finally, we \textbf{evaluate causal head replacement} using an evaluation suite comprising six benchmark datasets: HellaSwag \cite{Zellers2019}, PIQA \cite{Bisk2020}, SciQ \cite{Welbl2017}, ARC-Easy \cite{Clark2018}, Social IQA \cite{Sap2019}, and COPA.

\paragraph{Models}

We perform our analysis on four transformer architectures: BERT-base \cite{Devlin2019} containing 144 heads, GPT-2-small \cite{Radford2019} containing 144 heads, TinyLlama-1.1B \cite{Zhang2024} containing 704 heads, and Llama-3B \cite{Llama3Herd2024} containing 672 heads across 28 layers. Note that BERT-base is a \textit{bidirectional} attention model, while GPT-2-small, TinyLlama-1.1B, and Llama-3B are \textit{causal} attention models. 

\section{Experiments}

\subsection{Attention Alignment Analysis}
\label{sec:alignment_eval}

\begin{figure}[t]
  \centering
  \includegraphics[width=\textwidth]{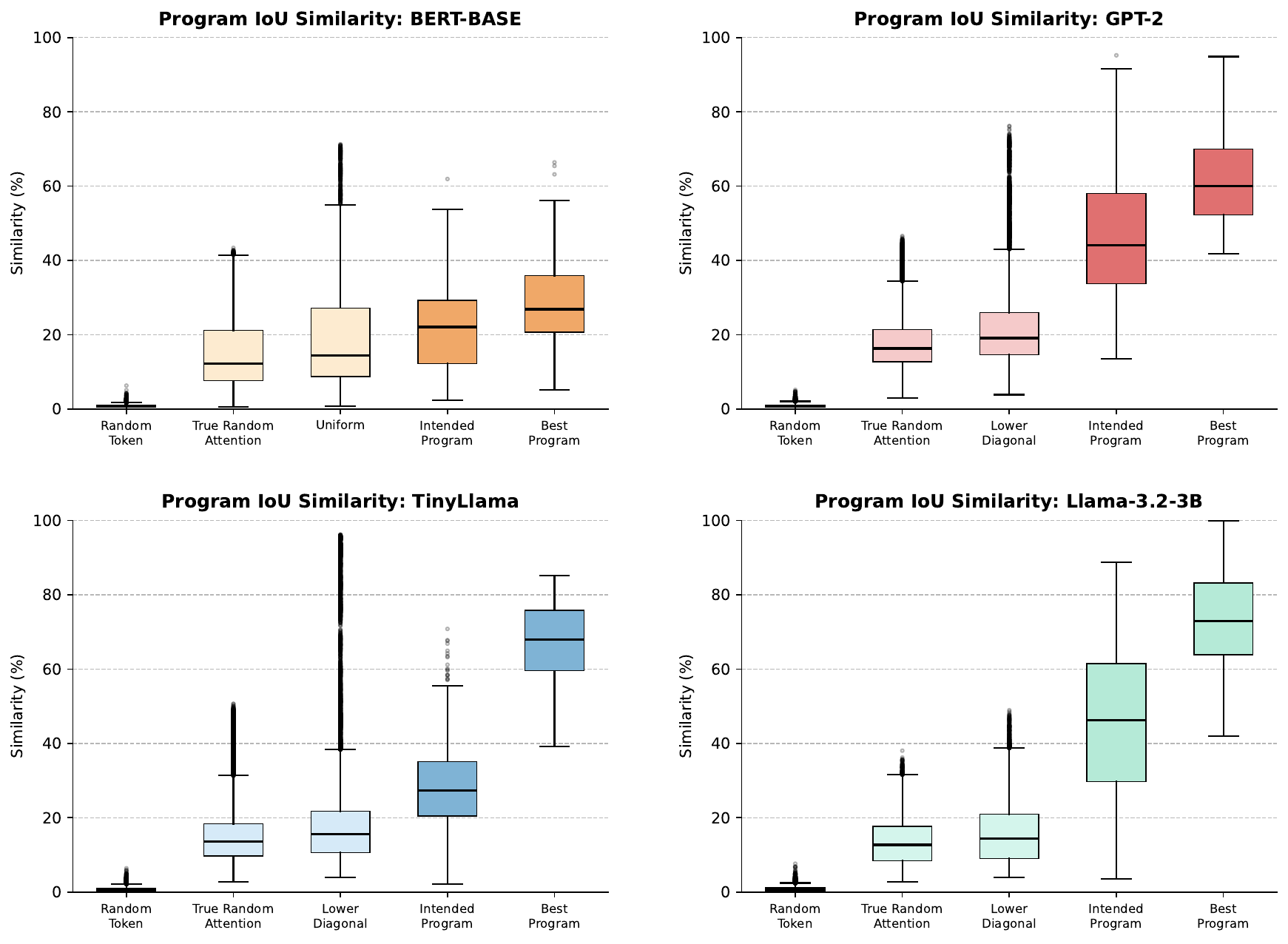}
  \caption{Analysis of program Intersection-over-Union similarity scores across all model attention heads. In general, heads in autoregressive models are easier to fit than in bidirectional models; fit quality increases with model scale. In many cases, the program synthesized for a given head (Intended Program) is outperformed by a program synthesized for a distinct head (Best Program).}
  \label{model-level-analysis}
\end{figure}

We evaluate the IoU similarity of predicted attention patterns via our synthesized programs to ground-truth attention patterns. We perform a comparative analysis summarized in Figure~\ref{model-level-analysis}, where we include two random baselines: \textbf{Random Token}, which assigns each query's full attention to a randomly selected token in each row (target token), and \textbf{Random Column}, which concentrates attention uniformly on a single column (source token) across all rows (target tokens). We additionally include a \textbf{Uniform Attention} baseline that assigns attention fully to the lower diagonal for the causal models and to all tokens equally in BERT. We compare these against two program-based conditions: the \emph{intended} program synthesized specifically for each head, and the \emph{best} program, defined as the highest-scoring selected by maximizing IoU across the library of all programs synthesized for all heads in a model. (The latter procedure is related to quality--diversity optimization algorithms like MAP-Elites \citep{mouret2015illuminating}.) 

We find that the best program for each head significantly outperforms random and uniform baselines across every model. The (globally) best program also consistently outperforms the ``intended program'' synthesized based on data for each individual heads, reflecting that some synthesized functions are sufficiently general to approximate multiple heads with high fidelity. Concretely, the best-fit program differs from the individually-synthesized (intended) program for 95.8\% of GPT-2 heads and 99.9\% of TinyLlama-1.1B heads, with the best-fit substitution improving mean IoU by 0.15 and 0.39 respectively over using the intended program alone (Appendix~\ref{sec:appendix_best_vs_intended}).
We further observe a systematic difference between encoder and decoder models, and a positive relationship between model scale and IoU similarity. BERT is comparatively poorly characterized by its synthesized programs relative to its decoder counterparts, which we attribute to the masked language modeling objective producing bidirectional attention distributions that are more complex and less amenable to symbolic approximation. Decoder models, by contrast, are constrained to attend only to preceding tokens, reducing the search space and making individual head behaviors more predictable. Notably, IoU scores also increase with model scale: GPT-2 achieves a mean best-program IoU of 69\%, TinyLlama-1.1B of 74\%, and Llama-3B of 79\%. We hypothesize this reflects functional specialization at scale, where a larger number of attention heads leads each individual head to serve a narrower and more symbolically tractable role.

\subsection{Qualitative Analysis of Best-Fit Heads}

\begin{figure}[t]
  \centering
  \begin{subfigure}[b]{0.47\linewidth}
    \centerline{\includegraphics[width=\linewidth]{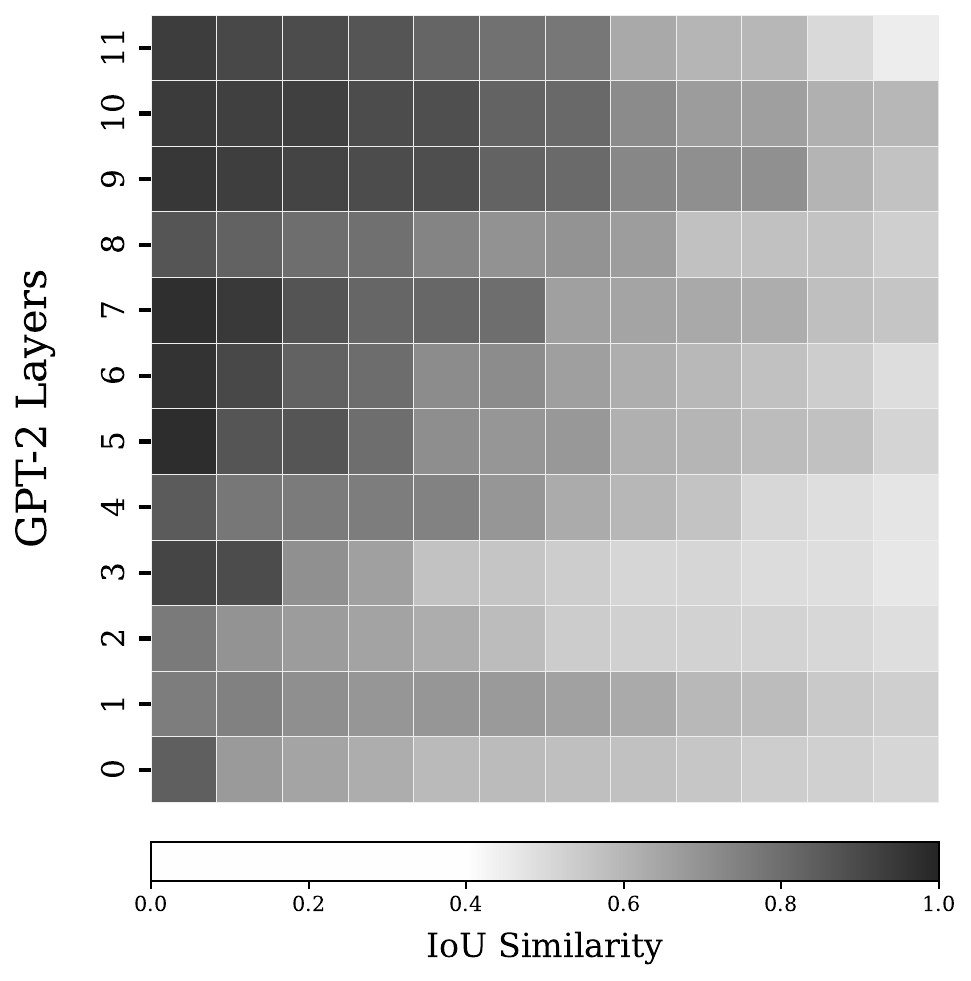}}
    \caption{
      GPT-2 Attention Head Score Heatmap
    }
    \label{figure_a}
  \end{subfigure}
  \hfill
  \begin{subfigure}[b]{0.5\linewidth}
    \centerline{\includegraphics[width=\linewidth]{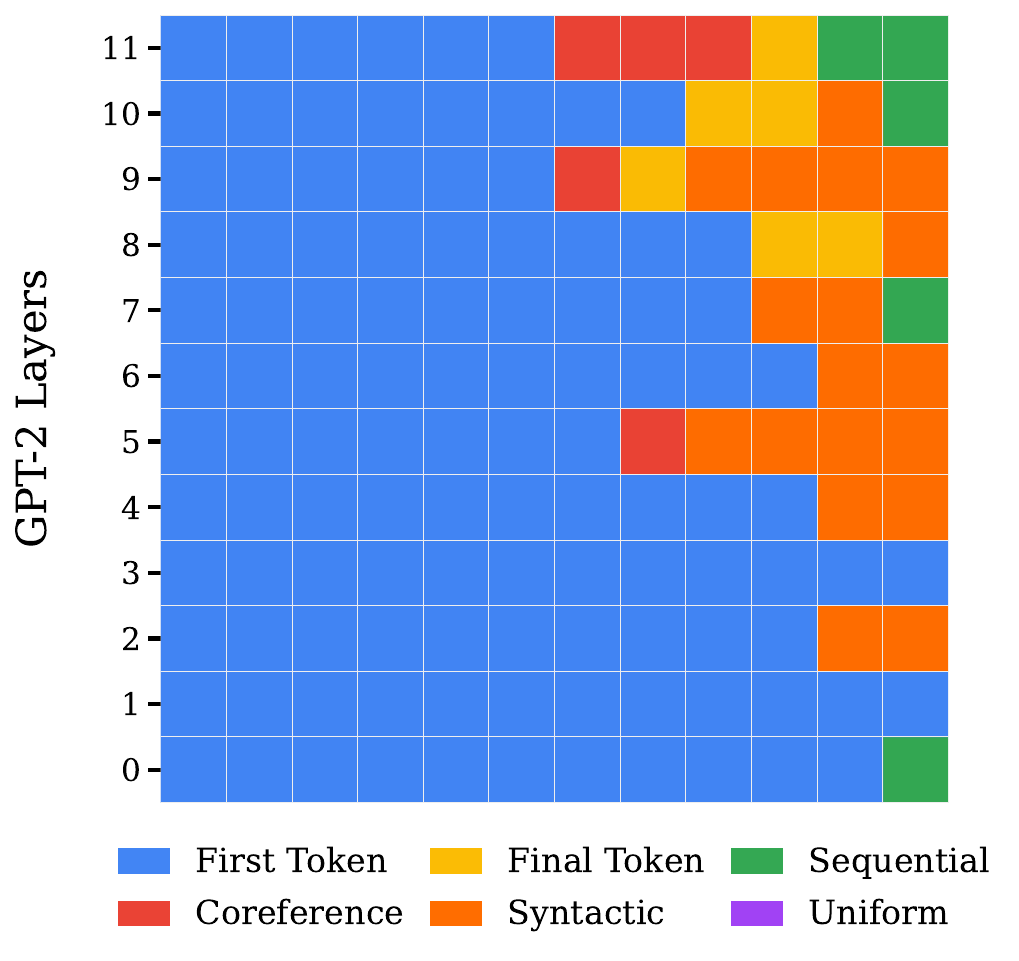}}
    \caption{
      GPT-2 Attention Head Category Outline
    }
    \label{figure_b}
  \end{subfigure}
  \caption{GPT-2 similarities and program types by layer. (a) Attention head accuracies (darker is more accurate). We sort the heads in each layer, and observe that earlier layers are generally harder to approximate than later ones. (b) Attention head types, automatically clustered based on program docstrings. Here we sort each row by type (so note that a given cell of (a) does not necessarily correspond to the same cell of (b)!). A substantial fraction of heads in GPT-2 are well approximated by programs that attend primarily to the first token (individual implementations may behave slightly differently on other tokens, so this coarse categorization may mask significant behavioral differences). Linguistic and discourse functions appearing deeper in the model (q.v.\ \citep{Tenney2019}).}
  \label{fig:full-model-outline}
\end{figure}

Figure ~\ref{figure_a} and ~\ref{figure_b} provide a head-level visualization of GPT-2 programs types and their corresponding fits, pairing (a) an IoU score heatmap with (b) a program category heatmap in which each head is colored by the functional category of its best-fit program. For (b), we constructed six categories of head types by prompting the synthesis agent to group the full set of program, providing a coarse decomposition of the symbolic logic present in the library.

From (a), we observe high alignment across the majority of heads, with later layers appearing to contain a greater number of heads that are easily explained by our set of programs. From (b), we see that program categories are not uniformly distributed across depth: first-token programs dominate early layers, while syntactic programs account for a larger share of middle-layer assignments. 

\subsection{Head Replacement Analysis}
\label{sec:replacement_eval}

\begin{figure}[t]
  \centering
  \begin{subfigure}[b]{0.49\linewidth}
    \centerline{\includegraphics[width=\linewidth]{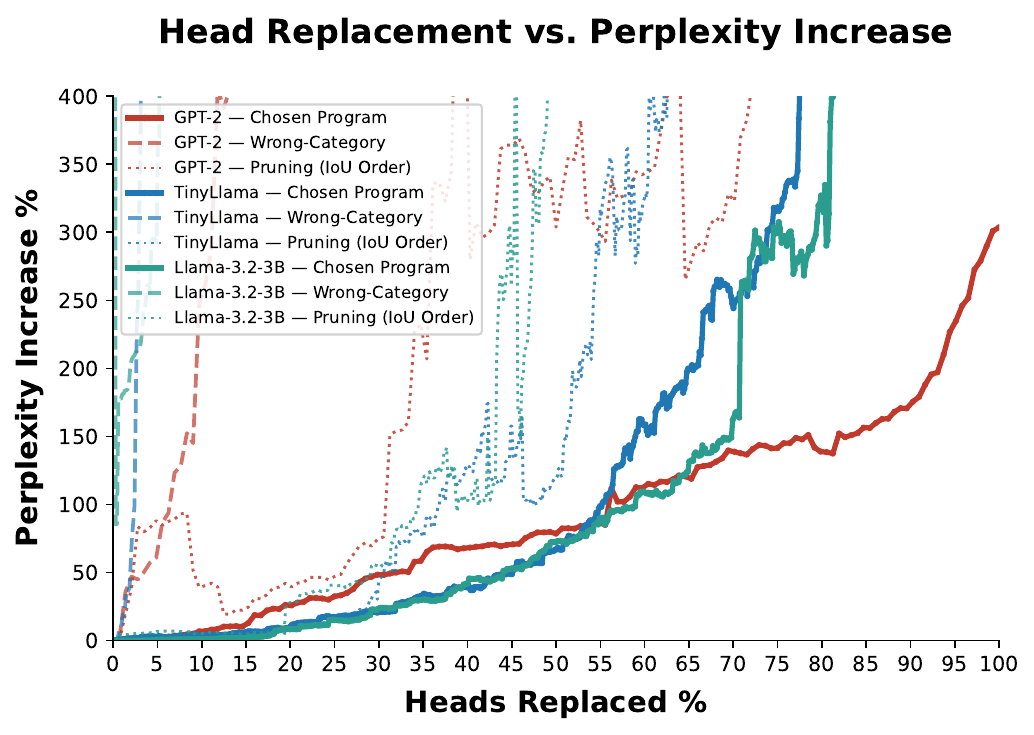}}
    \caption{Normalized Perplexity Increase (\%) versus the quantity of attention heads replaced by programs.}
    \label{perplexity-analysis}
  \end{subfigure}
  \hfill
  \begin{subfigure}[b]{0.49\linewidth}
    \centerline{\includegraphics[width=\linewidth]{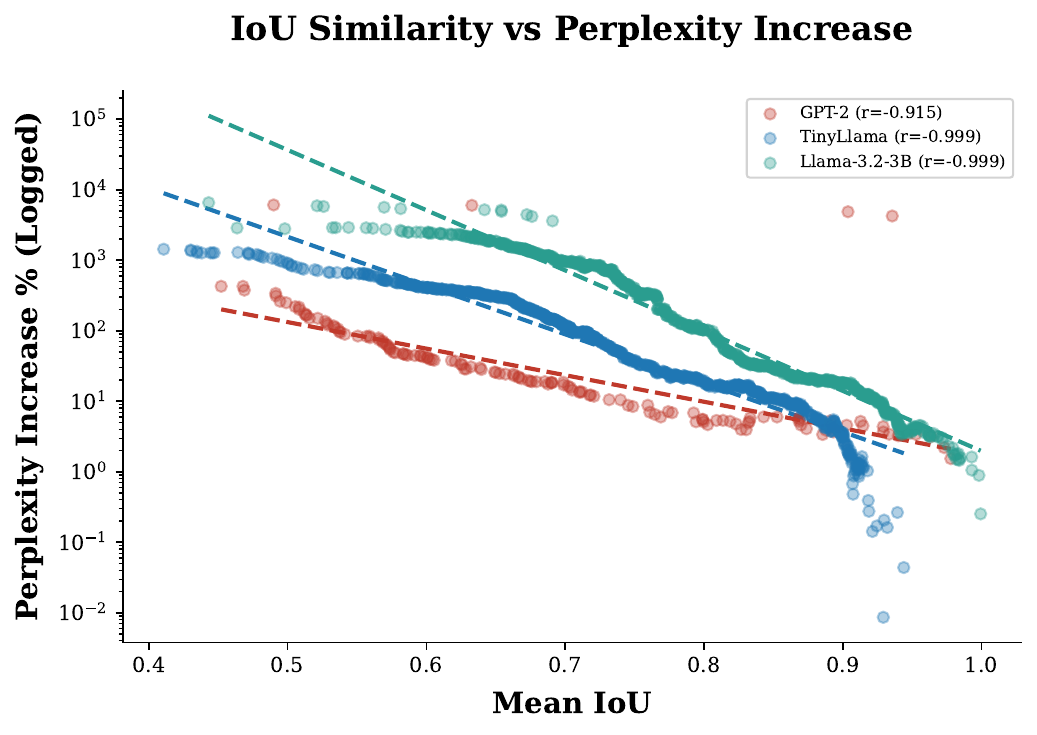}}
    \caption{Correlation of Normalized Perplexity Increase (\%) to IoU Similarity in Head and Program pairs}
    \label{correlation-score}
  \end{subfigure}
  \caption{Perplexity remains low when high-IoU heads are replaced first (left), consistent with the strong negative correlation between programmatic alignment and substitution cost across head-program pairs (right). We also plot a pruning baseline, which performs worse than programs, especially as more heads are replaced.}
  \label{fig:perplexity_and_correlation}
\end{figure}

To move beyond similarity metrics and explore causal evidence of functional alignment, we perform a series of interchange interventions where original attention patterns are replaced by the outputs of our symbolic proxies. Heads are replaced greedily in descending order of IoU score, representing a replacement trajectory in which the best-fit heads are substituted first. 

We begin by examining perplexity changes on our dataset processed by models with original heads and our programmatic proxies. Because this dataset was the same one we used to derive our programmatic attention approximations, this evaluation represents an \textit{in-distribution} evaluation. Our experimental setup tracks the Normalized Perplexity Increase (\%) on the y-axis against the percentage of heads replaced on the x-axis. Normalization is employed as perplexity values derived from different datasets and model states exist in distinct distributions; by calculating only the increase from the unedited model, we isolate the impact of the intervention itself. 

\begin{figure}[h]
  \centering
  \includegraphics[width=\textwidth]{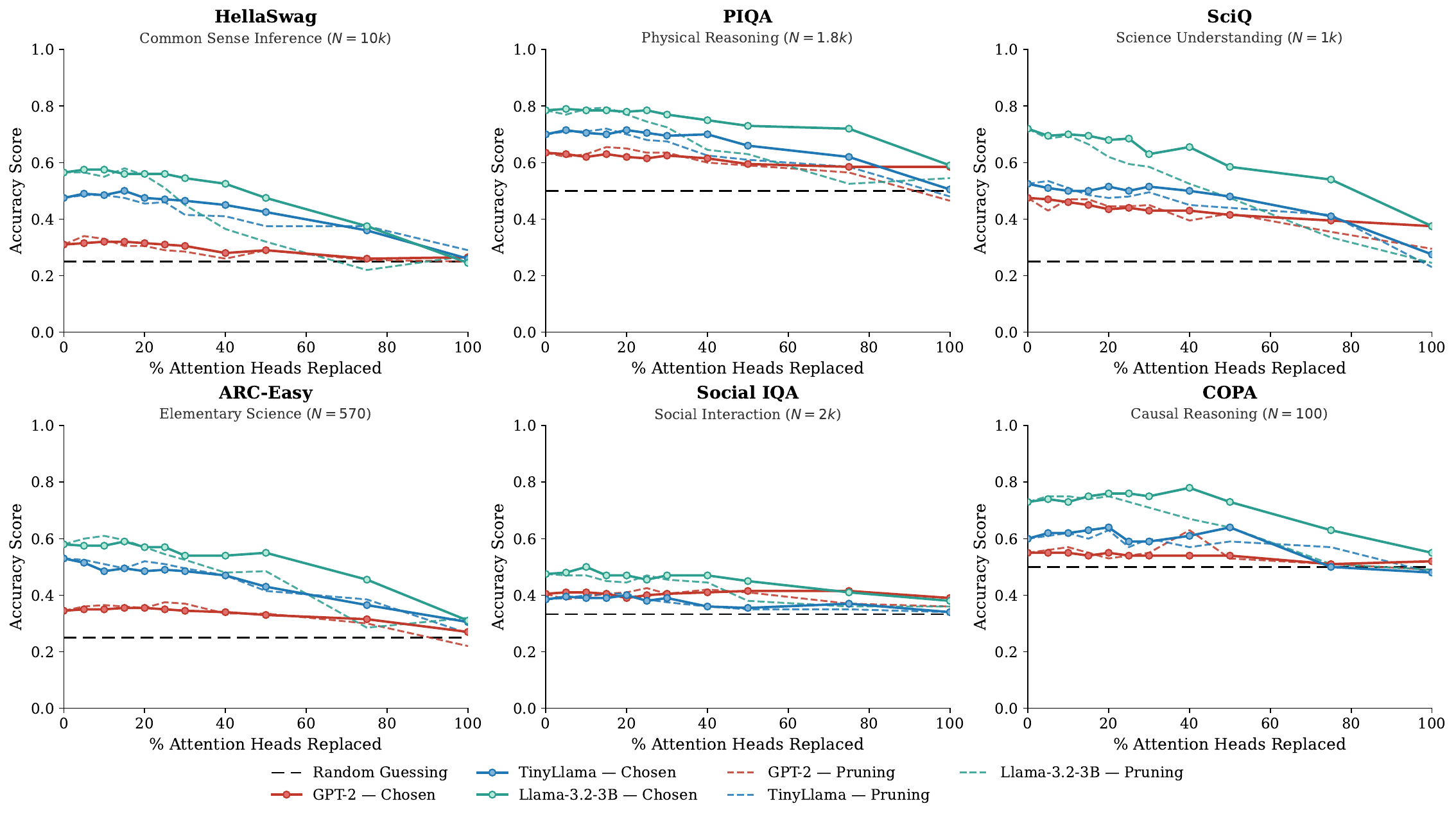}
  \caption{Effect of replacing attention heads on downstream model evaluations. We find that replacing heads with their best fit programs (solid) performs better on downstream tasks compared to pruning those same heads in IOU descending order (dashed). Interestingly, there is substantial overlap between the two conditions in GPT-2 and TinyLlama-1.1B, where the curves are nearly indistinguishable until roughly half and a third of heads are replaced respectively; Read across three models, the pattern is that  for the most removable heads, substituting a program and deleting the head outright are interchangeable, and programs pull ahead once replacement reaches heads that are doing work the benchmarks can detect. }
  \label{replacement-evaluation}
\end{figure}

Figure~\ref{perplexity-analysis} illustrates the normalized perplexity increase across models when original attention patterns are replaced by our synthesized programs and by two ablations applied in the same replacement order. In the \textbf{wrong-category} ablation, each replaced head receives a program drawn from a functional category other than its own; in the \textbf{zero-ablation}, the head's contribution is set to zero outright. Wrong-category substitution increases above 1000\% within the first 5--10\% of heads, while the degradation due to program replacement is slower. While zero-ablation is less disruptive than wrong-category substitution, it is nonetheless more disruptive than program substitution GPT-2 and Llama-3.2-3B at every level we test. In TinyLlama-1.1B, however, the two conditions are indistinguishable below roughly a quarter of heads, and programs pull ahead only above that point. We read this as evidence that the earliest heads in IoU order are largely attention sinks whose contribution is close to removable, and that the programs' advantage comes from the heads replaced after them.

Next, we evaluate the \textit{out-of-distribution generalization} of our symbolic heads to a wide range of multiple-choice natural language reasoning tasks. Figure~\ref{replacement-evaluation} summarizes the performance of models across six tasks spanning science, physical, commonsense, and causal reasoning. Our results demonstrate that programmatic substitution of up to 30--40\% of attention heads does not significantly degrade downstream task performance, a finding that holds across all four architectures tested. Even at higher replacement levels, the performance on several benchmarks remains non-trivially better than random. 

\section{Related Work}

\paragraph{Interpreting Deep Networks}

Transformer language models can be interpreted at several levels of granularity. At the neuron level, individual units can be examined for the information they encode \cite{dalvi2019one}, offering granular insight but limited generalizability \cite{Cunningham2023}. At the model level, analyses consider how layers jointly shape predictions \citep{Tenney2019}, revealing high-level strategies while obscuring specific circuits \cite{Bricken2023}. At the attention-head level, one can analyze how heads distribute focus across a sequence \cite{Vig2019}; this balances interpretability and simplicity, since heads often track linguistic patterns that can be assigned definitive roles.

\paragraph{Explaining Attention}

Attention weights determine how the contents of each token in one layer map onto the next.
Prior work treats these weights as an \textit{attribution map}: examining where a layer attends gives a rough sense of which tokens matter most to the model's prediction \cite{JainWallace2019}, and tracking attention across layers offers a signal for the circuits and internal mechanisms in use \cite{Vig2019, Elhage2021}. Whether attention weights themselves constitute faithful explanations remains debated \citep{JainWallace2019, WiegreffePinter2019}; our approach sidesteps this question by evaluating symbolic proxies \emph{causally}, through direct replacement and measurement of downstream behavior, rather than treating attention weights as explanations in themselves.

Attention frequently reflects structured, functionally specialized behavior: certain BERT heads consistently track specific syntactic relationships such as subject-verb agreement and direct objects \cite{ClarkEtAl2019}, while others remain diffuse or task-agnostic. Some heads are also critical for integrating information during in-context learning, as removing them impairs model adaptation \cite{YinSteinhardt2025}.

A related line of work studies \emph{pruning} attention heads \citep[e.g.][]{voita2019analyzing, Michel2019}, finding that many can be removed without substantially harming performance. We instead focus on explaining all heads---even those unnecessary for prediction---since such heads may still capture features that predict generalization \citep{li2025can}.

\paragraph{Interpretability and Program Synthesis}

Closely related to our work, \cite{Mu2020} approximate neuron behavior in vision models and small transformer encoders via enumerative search over simple logical expressions. A parallel line of automated-interpretability work uses LMs to generate natural-language descriptions of neurons and features \cite{Bills2023}, but such descriptions lack formal verifiability and cannot be substituted into model computations. Other work builds transformers that are interpretable by design, either by compiling human-written RASP programs into weights \cite{Weiss2021} or training modified transformers that convert into discrete, human-readable programs \cite{Friedman2023}; these approaches are formally precise but require architectural constraints or training from scratch, rather than post-hoc analysis of pretrained models. \cite{Michaud2024} convert RNNs trained on algorithmic tasks into finite state machines and apply symbolic regression to distill Python code, identifying generalization to transformers and larger networks as open directions.

Our method extends this line of work by applying LM-guided program synthesis \cite{Austin2021, Chen2021Codex, Olausson2023} directly to attention heads in modern autoregressive LMs, targeting the richer space of Python programs over compact boolean or integer expressions. Where past work was limited to simple expressions on small or specialized architectures, we scale to GPT-2, TinyLlama-1.1B, and Llama-3B, and validate our approximations causally by inserting synthesized programs into live forward passes---to our knowledge the first programmatic explanations to scale to modern LMs, and the first evidence that such programs preserve model capabilities when substituted in.

\section{Discussion}

The results presented in this work demonstrate that a symbolic approach to mechanistic interpretability is not only feasible but could yield direct proxies for complex neural activations. Model-level analyses in Figure \ref{model-level-analysis} show that executable programs can achieve up to 99\% mean IoU similarity against observed attention patterns. Our evaluation in Figure \ref{fig:perplexity_and_correlation} also demonstrates that there is a clear negative correlation between the mean IoU similarity of a program and head  and the perplexity increase as a result of replacing the head with its program. Examining the model's performance trends on downstream question-answering evaluations, we find that we can replace up to 30-40\% of all attention heads with their highest-similarity programs $\pi$ without losing task ability. This indicates that our method can almost completely capture the functional role of up to 30-40\% of all LM heads.

\paragraph{Limitations} There is significant room for improvement in expanding the diversity and complexity of these hypotheses. Currently, no model's attention heads are fully characterized: a substantial fraction of heads achieve IoU scores below 60\%, and we attribute the question-answering degradation observed at high replacement levels in Figure~\ref{replacement-evaluation} primarily to these poorly fit heads. We further observe that many high-scoring programs are not particularly complex, and that the improvement in question-answering seen for some models at low replacement levels may be akin to a pruning effect. Closing this gap likely requires both richer synthesis strategies such as multi-round refinement with stronger feedback signals. The goal of this paper is to establish that a small, curated program library can serve as faithful proxies for real attention heads with few refinements, providing a concrete foundation for future work toward complete symbolic characterization of transformer architectures.

\paragraph{Next steps} One objective for subsequent research is to achieve a \textit{complete} symbolic characterization of LMs. By bridging the gap between neural activations and symbolic code, we aim to demonstrate that complex model behaviors can be distilled into human-readable logic, allowing us to completely trace the allocation of attention throughout the entire architecture. Ultimately, a complete symbolic characterization of a LM would allow researchers to reason about model behavior the way they reason about algorithms: by reading, modifying, and testing the underlying logic directly.




\subsection*{Impact Statement}

We do not anticipate any harms or misuses associated with the methods described in this paper.

\bibliography{example_paper}
\bibliographystyle{plain}

\newpage
\appendix

\section{Appendix}

To evaluate the breadth of our synthesized program library $\Pi$, we perform a model-wide alignment analysis across all four architectures. For every attention head in each model, we identify the highest-fidelity programmatic proxy within our library and report its maximum alignment score $S = \text{IoU}(A, \hat{A})$.

\subsection{BERT-base}

\begin{figure}[H]
  \centering
  \begin{subfigure}[b]{0.47\linewidth}
    \centerline{\includegraphics[width=\linewidth]{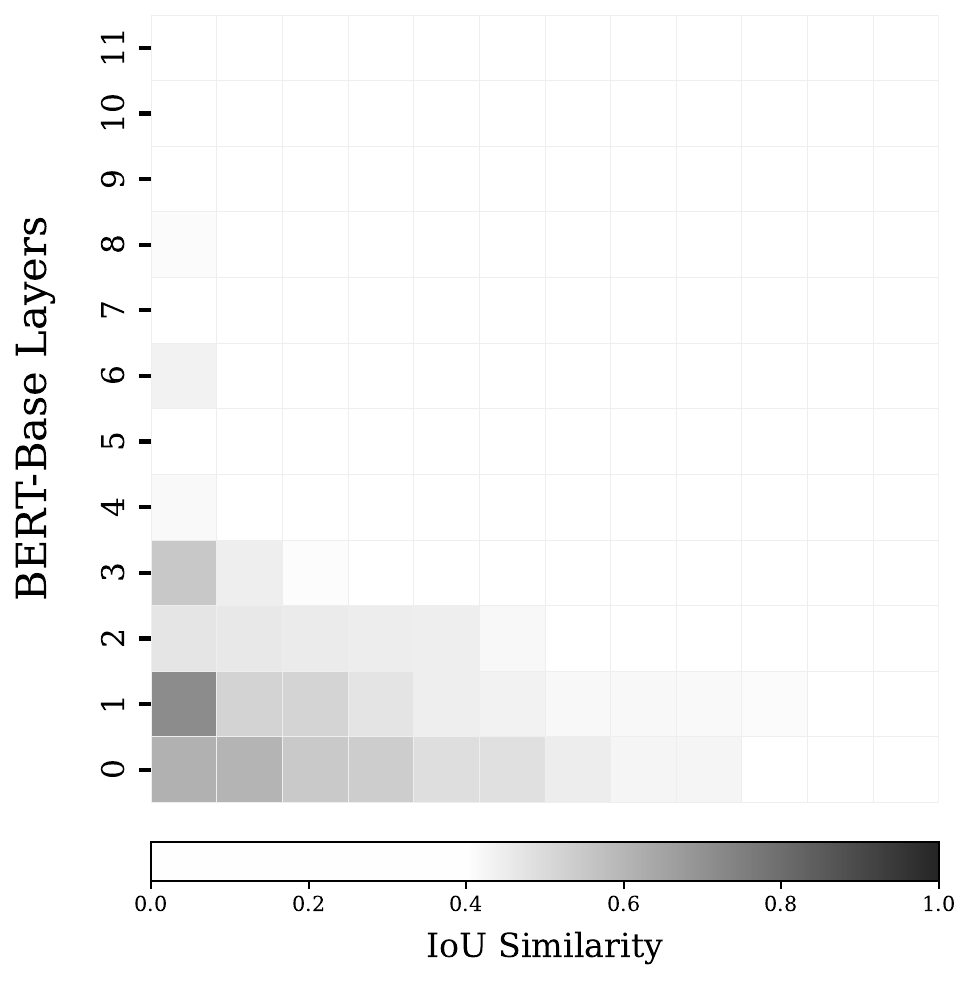}}
    \caption{BERT-base IoU score heatmap. Each cell represents the maximum IoU score achieved by a programmatic proxy for that head.}
    \label{fig:bert-score}
  \end{subfigure}
  \hfill
  \begin{subfigure}[b]{0.5\linewidth}
    \centerline{\includegraphics[width=\linewidth]{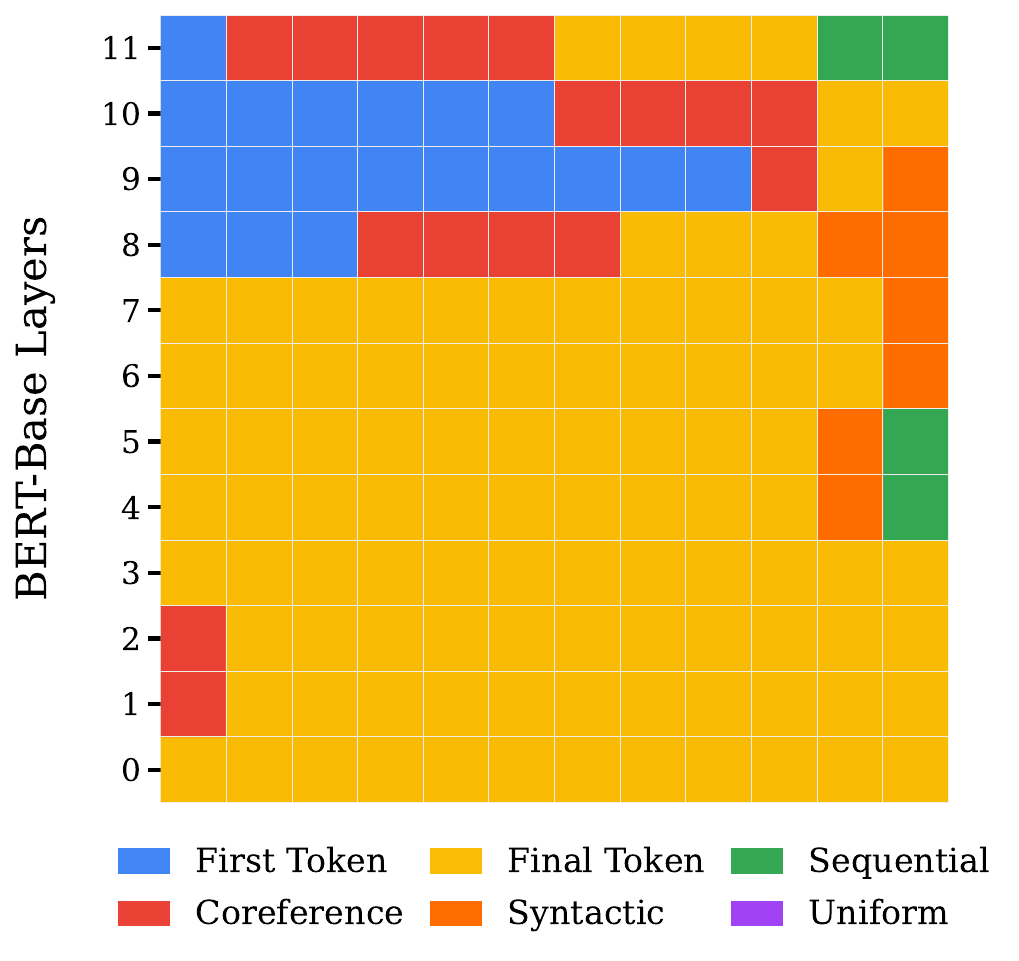}}
    \caption{BERT-base program category heatmap. Circle color indicates the category of the best-fit program for each head.}
    \label{fig:bert-cat}
  \end{subfigure}
  \caption{Head-to-program alignment for BERT-base. Dark cells indicate heads whose behaviors are not yet well-captured by the current program library.}
  \label{fig:bert-heatmap}
\end{figure}

Figure~\ref{fig:bert-heatmap} shows that BERT-base is the most poorly characterized model in our analysis, with the majority of heads achieving low single-program IoU scores. High-fidelity matches are sparse and isolated rather than layer-wide, suggesting that the current program library does not yet capture the dominant functional logic of encoder attention. We hypothesize this reflects the inherent difficulty of the masked language modeling objective: compressing bidirectional contextual information into compact attention distributions likely demands more diverse and abstract programs than autoregressive next-token prediction. The scattered high-similarity heads that do exist tend to align with positional and linguistic programs, confirming that these primitives are architecture-agnostic, but the overall coverage gap for BERT represents the clearest direction for future library expansion \citep{ClarkEtAl2019}.

\subsection{TinyLlama-1.1B}

\begin{figure}[h]
  \centering
  \begin{subfigure}[b]{0.475\linewidth}
    \centerline{\includegraphics[width=\linewidth]{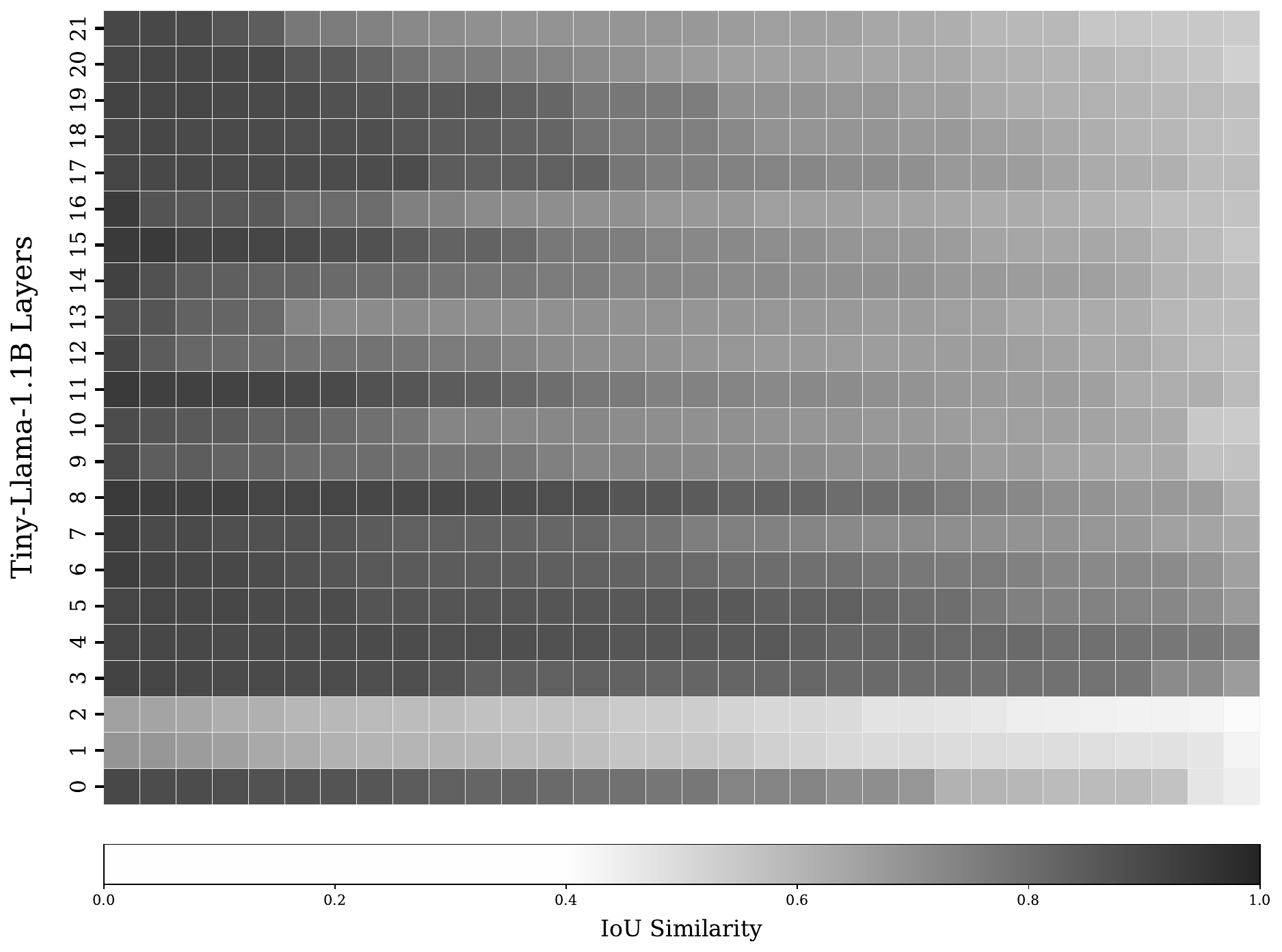}}
    \caption{TinyLlama-1.1B IoU score heatmap.}
    \label{fig:tinyllama-score}
  \end{subfigure}
  \hfill
  \begin{subfigure}[b]{0.49\linewidth}
    \centerline{\includegraphics[width=\linewidth]{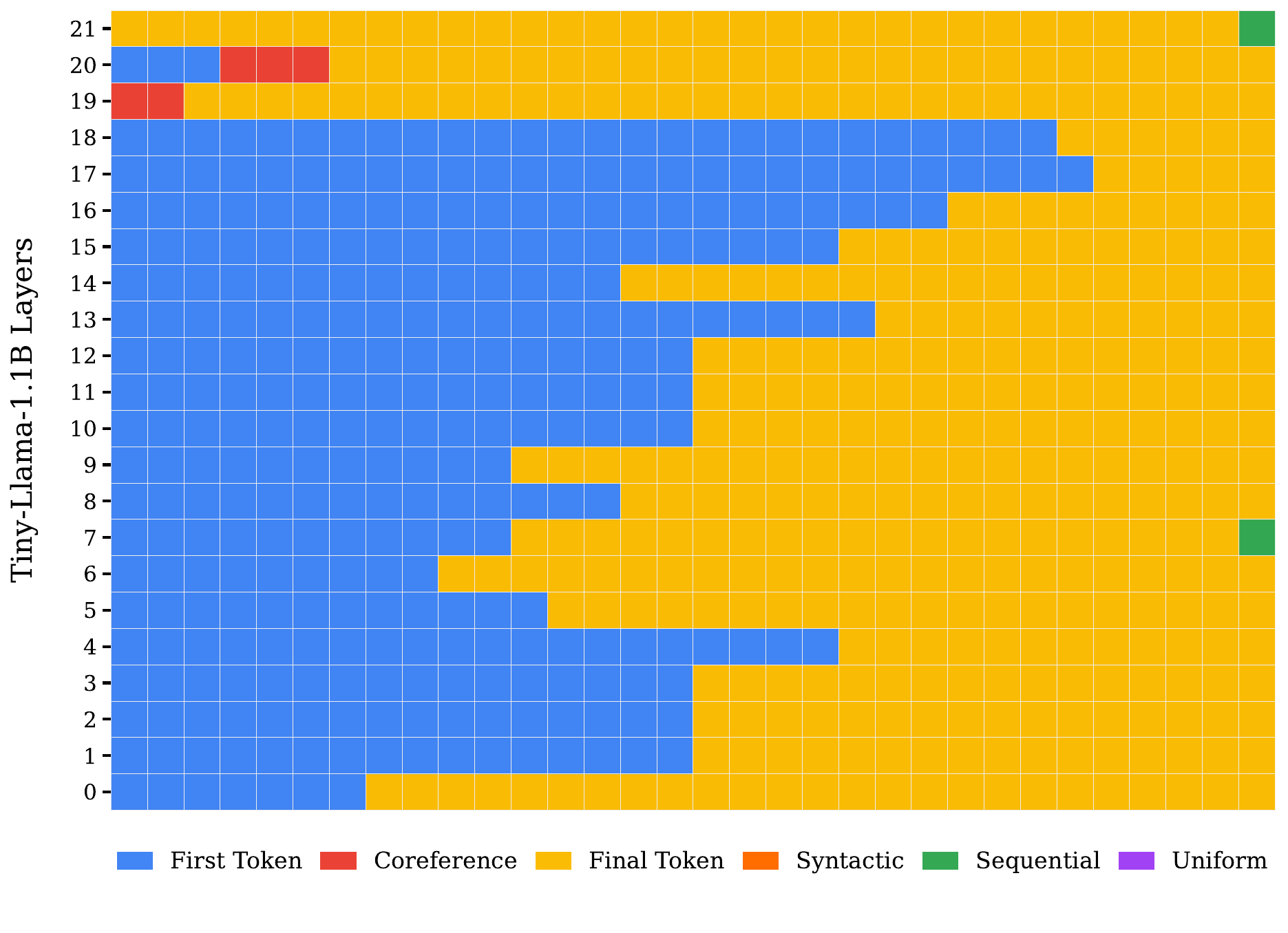}}
    \caption{TinyLlama-1.1B program category heatmap.}
    \label{fig:tinyllama-cat}
  \end{subfigure}
  \caption{Head-to-program alignment for TinyLlama-1.1B across 22 layers and 32 heads per layer.}
  \label{fig:tinyllama-heatmap}
\end{figure}

TinyLlama exhibits a pattern broadly similar to GPT-2, with initiation and anchoring programs dominating early layers and more varied program assignments at greater depth. The larger head count per layer (32 heads versus 12 in GPT-2) reveals finer-grained functional specialization, with distinct program categories occupying well-defined layer bands. The near-uniform lower-diagonal structure in many heads reflects the strong positional prior imposed by causal language modeling at this scale.

\subsection{Llama-3.2-3B}

\begin{figure}[h]
  \centering
  \begin{subfigure}[b]{0.47\linewidth}
    \centerline{\includegraphics[width=\linewidth]{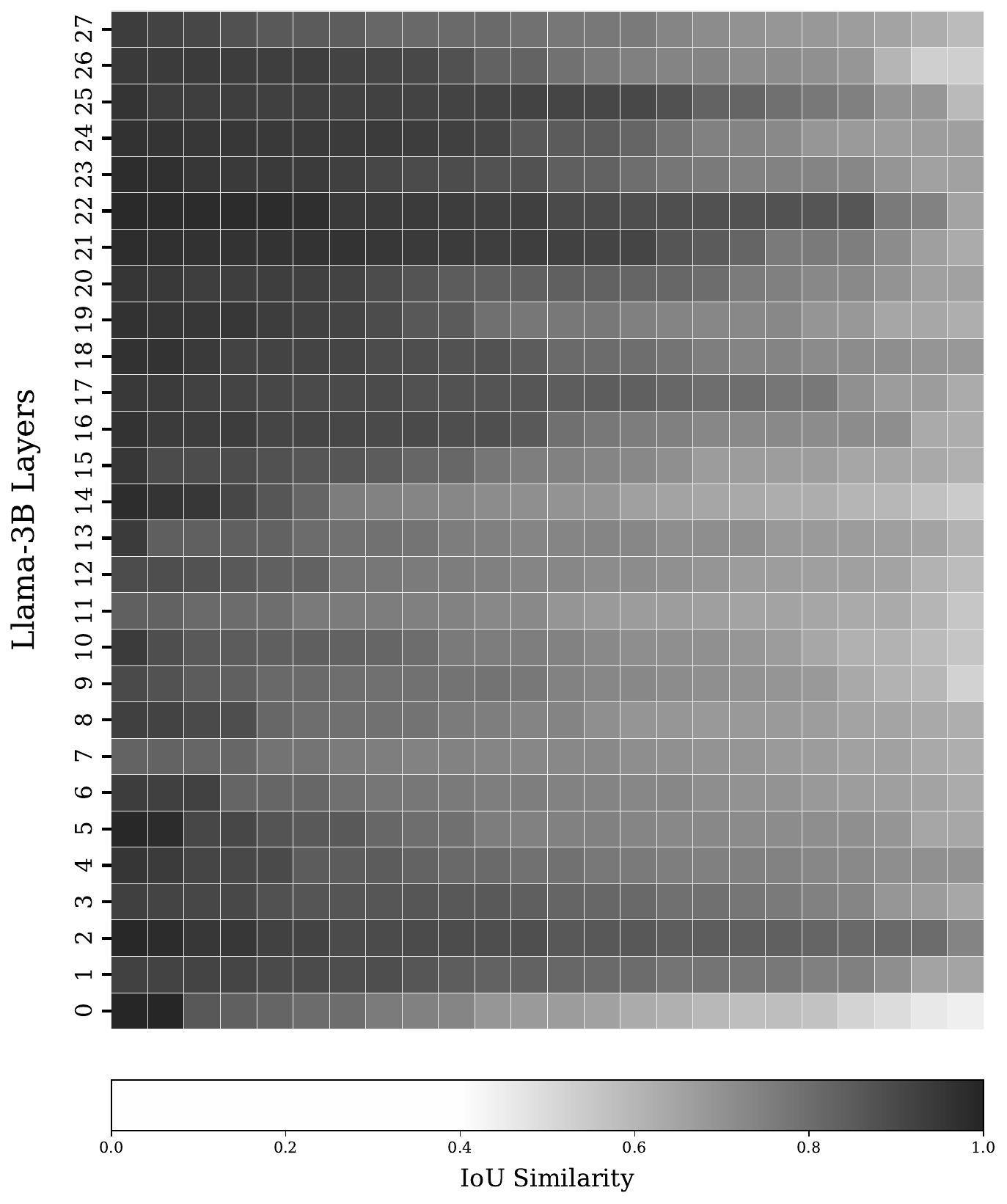}}
    \caption{Llama-3.2-3B IoU score heatmap. Dark regions indicate heads whose behaviors are not yet well-captured by the current program library. While the plot shows many heads as being drawn from a small number of clusters, there are substantial differences in behavior within each cluster (e.g.\ many tokens that primarily attend to the first token, but occasionally implement some other function).}
    \label{fig:llama-score}
  \end{subfigure}
  \hfill
  \begin{subfigure}[b]{0.48\linewidth}
    \centerline{\includegraphics[width=\linewidth]{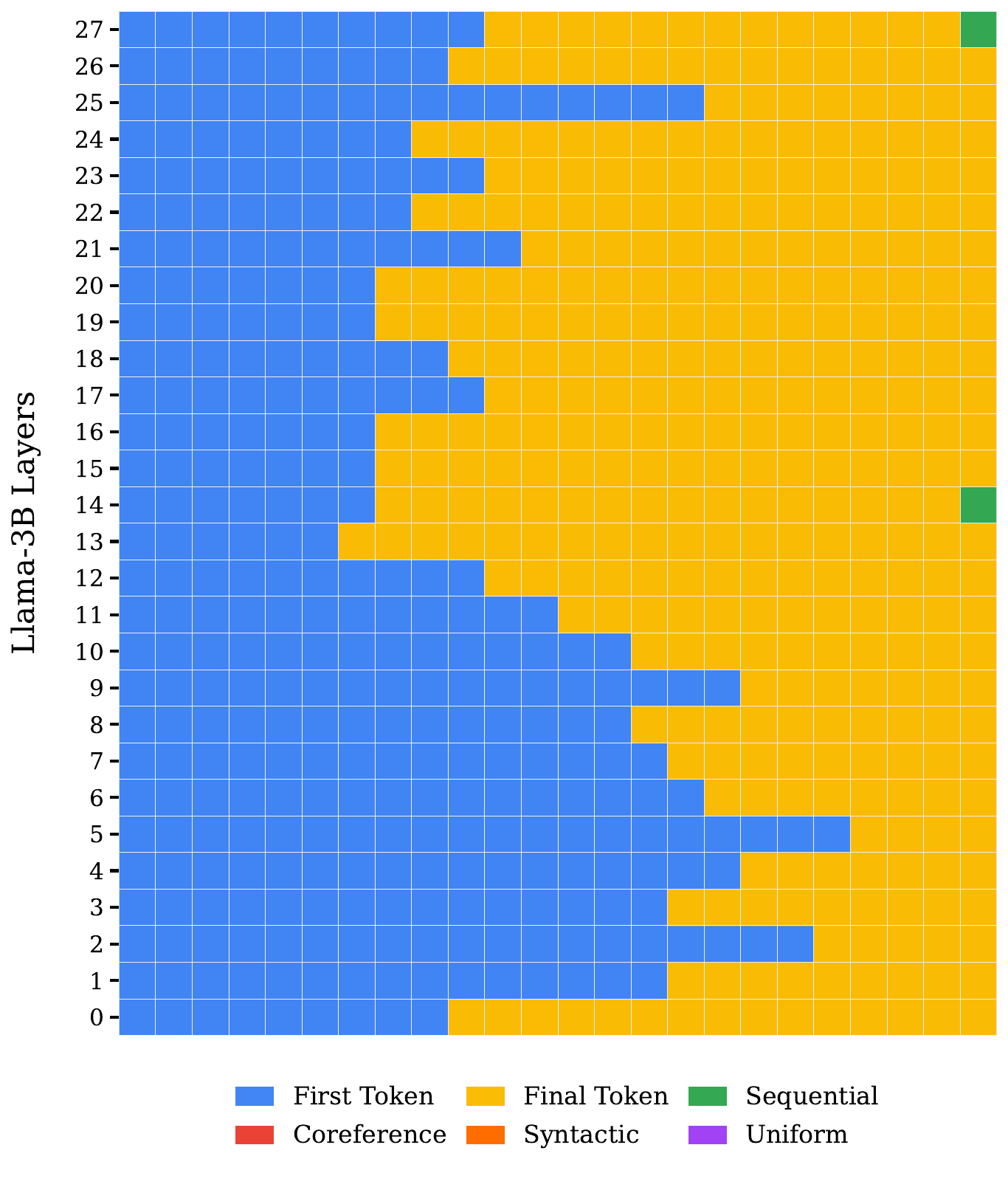}}
    \caption{Llama-3.2-3B program category heatmap. Coverage gaps are distributed throughout model depth rather than localized to specific layers.}
    \label{fig:llama-cat}
  \end{subfigure}
  \caption{Head-to-program alignment for Llama-3.2-3B across 28 layers and 24 heads per layer. }
  \label{fig:llama-heatmap}
\end{figure}

Llama-3B is dominated by a narrow set of recurring programs across nearly all layers and heads, but with generally moderate IoU scores rather than the high-confidence matches seen in GPT-2. This pattern suggests that the current library identifies the correct functional family for most heads but lacks the resolution to capture the specific variants operating at this scale. Isolated dark cells appear throughout the depth of the model rather than concentrating in any particular region, indicating that the coverage gap is distributed rather than localized to specific functional stages. Across all four architectures, the same foundational positional and anchoring primitives appear in the earliest layers, supporting the view that basic information-routing behaviors are largely invariant to model scale and that complexity accumulates in how those primitives are extended and combined at greater depth.

\subsection{Synthesis Agent Prompt}
\label{sec:appendix_prompt}

Program synthesis was performed using Claude Sonnet 4 and GPT-4o as synthesis agents $\mysynthesizer$. Each head was queried with a prompt summarizing its empirical attention pattern (stratified by attention-weight percentile across several example sentences) together with the synthesis task specification and available helper functions. The full prompt-construction code and helper-function implementations will be included in our public code release. An illustrative excerpt, abridged to omit a subset of example sentences and helper-function definitions, is shown below:

\begin{quote}
\small
\begin{verbatim}
I'm analyzing attention head L0H1 of GPT2-small. Below are examples of
its attention patterns on several sentences. Each example shows, for
each query token, a sample of key tokens it attends to with their
attention weights.

Example 1:
Sentence: "He followed me everywhere!" Lily said with a smile."

high attention (p75-p100, 0.999-1.000):
  'He'[0] -> 'He'[0] (1.000)
  ' everywhere'[3] -> ' everywhere'[3] (1.000)
  ' me'[2] -> ' me'[2] (0.999)
  (+ 1 more self-attention edges)
upper-mid attention (p50-p75, 0.993-0.999):
  ' followed'[1] -> ' followed'[1] (0.998)
  '!"'[4] -> '!"'[4] (0.998)
  ' smile'[9] -> ' smile'[9] (0.993)
lower-mid attention (p25-p50, 0.975-0.993):
  ' with'[7] -> ' with'[7] (0.988)
  ' a'[8] -> ' a'[8] (0.984)
  ' said'[6] -> ' said'[6] (0.975)
low attention (p0-p25, 0.014-0.975):
  '.'[10] -> '.'[10] (0.968)
  '.'[10] -> '!"'[4] (0.022)
  ' said'[6] -> ' me'[2] (0.014)

[...]

## Your task

Write a Python function `predict_attention_map(sentence: str) ->
tuple[list[str], np.ndarray]` that approximates this attention head's
behavior.

The function should:
1. Take a raw sentence string as input
2. Return a tuple of (tokens, attention_matrix) where:
   - `tokens` is a list of GPT2 BPE token strings (use the
     `gpt2_tokenize` helper)
   - `attention_matrix` is an (n, n) numpy array where n = len(tokens)
   - Each row should sum to 1.0 (row-stochastic)
   - The matrix must be lower-triangular (causal mask): token i can
     only attend to tokens j <= i

## Available helper functions

The following helper functions are available in your execution
environment (do NOT redefine them):

[...]

## Instructions

- Analyze the attention patterns carefully. Look for linguistic
  patterns like:
  - Positional patterns (attend to previous token, first token,
    self, etc.)
  - Syntactic patterns (attend to head of phrase, modifiers, etc.)
  - Mixed patterns (some positional bias + syntactic component)
- Use `gpt2_tokenize()` to get tokens (NOT spacy tokenization)
- Use `align_gpt2_to_spacy()` or `align_spacy_to_gpt2()` if you need
  linguistic features from spacy
- Use `make_row_stochastic()` and `apply_causal_mask()` to finalize
  the matrix
- Your function will be called on new sentences not shown above

Write ONLY the function definition inside a single ```python code
block. Do not include import statements for numpy or the helpers
(they are pre-imported). Do not include test code or example usage.
\end{verbatim}
\end{quote}

\subsection{Reproducibility Details}
\label{sec:appendix_repro}

The replacement order for the \emph{chosen} condition (Figures~\ref{perplexity-analysis} and \ref{replacement-evaluation}) is deterministic: heads are substituted in strictly descending order of IoU score, so there is no seed variation to report for that condition. For the \emph{wrong-category} baseline, each replaced head is assigned a uniformly random category (excluding its own best-fit category) and a uniformly random program within that category; we report results averaged over $N{=}3$ random seeds per (model, replacement level) pair in Table~\ref{tab:wrong_category_error_bars}. All downstream accuracies are computed over 200 held-out examples per benchmark, across 10 replacement levels ($5\%$--$100\%$ of heads). We note that our category-separation sufficiency check (maximum cross-category average-pattern IoU below a $0.4$ threshold) does not currently pass for any of the three models, so these error bars should be read as a measure of run-to-run stability under this randomization scheme rather than confirmation that wrong-category substitutions are drawn from a cleanly disjoint attention-pattern class.

\begin{table}[h]
\centering
\small
\begin{tabular}{llcccccc}
\toprule
Level & Model & HellaSwag & PIQA & SciQ & ARC-Easy & Social IQA & COPA \\
\midrule
25\% & GPT-2 & 0.268$\pm$0.026 & 0.563$\pm$0.014 & 0.385$\pm$0.017 & 0.310$\pm$0.031 & 0.370$\pm$0.010 & 0.527$\pm$0.072 \\
25\% & TinyLlama-1.1B & 0.287$\pm$0.023 & 0.480$\pm$0.026 & 0.255$\pm$0.040 & 0.255$\pm$0.058 & 0.362$\pm$0.023 & 0.553$\pm$0.012 \\
25\% & Llama-3B & 0.277$\pm$0.025 & 0.483$\pm$0.032 & 0.323$\pm$0.006 & 0.257$\pm$0.023 & 0.368$\pm$0.014 & 0.523$\pm$0.051 \\
\midrule
50\% & GPT-2 & 0.232$\pm$0.012 & 0.508$\pm$0.015 & 0.363$\pm$0.019 & 0.322$\pm$0.008 & 0.362$\pm$0.013 & 0.560$\pm$0.030 \\
50\% & TinyLlama-1.1B & 0.248$\pm$0.015 & 0.518$\pm$0.008 & 0.242$\pm$0.014 & 0.277$\pm$0.031 & 0.318$\pm$0.050 & 0.470$\pm$0.060 \\
50\% & Llama-3B & 0.260$\pm$0.035 & 0.492$\pm$0.025 & 0.302$\pm$0.021 & 0.238$\pm$0.006 & 0.332$\pm$0.048 & 0.520$\pm$0.053 \\
\bottomrule
\end{tabular}
\caption{Wrong-category baseline downstream accuracy (mean $\pm$ std over $N{=}3$ random seeds) at 25\% and 50\% of heads replaced. \emph{Chosen}-condition results in the main text are single-run since replacement order is deterministic; these error bars instead characterize run-to-run variability under matched-magnitude random head substitution.}
\label{tab:wrong_category_error_bars}
\end{table}

\subsection{Best-Fit vs. Intended Program}
\label{sec:appendix_best_vs_intended}

\begin{center}
\begin{tabular}{lccc}
\toprule
Model & \% heads best $\neq$ intended & Mean intended IoU & Mean best IoU \\
\midrule
GPT-2 & 95.8\% & 0.46 & 0.61 \\
TinyLlama-1.1B & 99.9\% & 0.28 & 0.67 \\
\bottomrule
\end{tabular}
\end{center}

The individually-synthesized (``intended'') program for a head is frequently outperformed by a program synthesized for a different head; \S\ref{sec:alignment_eval} discusses this in the context of quality--diversity search over the program library. We report this comparison for GPT-2 and TinyLlama-1.1B, for which a per-head intended-program baseline was generated; the same analysis was not run for Llama-3B.

\end{document}